\documentclass[lettersize,journal]{IEEEtran}
\usepackage{amsmath,amsfonts}
\usepackage{algorithmic}
\usepackage{array}
\usepackage{subfigure}
\usepackage{textcomp}
\usepackage{stfloats}
\usepackage{url}
\usepackage{verbatim}
\usepackage{graphicx}
\usepackage{caption}
\usepackage{color}
\usepackage{booktabs}
\usepackage{multirow}
\def\BibTeX{{\rm B\kern-.05em{\sc i\kern-.025em b}\kern-.08em
		T\kern-.1667em\lower.7ex\hbox{E}\kern-.125emX}}
\usepackage{balance}

\begin{document}
	
\title{Accurate and Real-time Pseudo Lidar Detection: Is Stereo Neural Network Really Necessary?}

\author{
	\IEEEauthorblockN{Haitao Meng\IEEEauthorrefmark{1}, Changcai Li\IEEEauthorrefmark{2}, Gang Chen\IEEEauthorrefmark{3}, Alois Knoll\IEEEauthorrefmark{1}} \\ \IEEEauthorblockA{\IEEEauthorrefmark{1}Technical University of Munich, Germany} \\
	\IEEEauthorblockA{\IEEEauthorrefmark{2}Guangxi Univeristy, China} \\
	\IEEEauthorblockA{\IEEEauthorrefmark{3} Sun Yat-sen University, China}
}

\maketitle

\begin{abstract}
The proposal of Pseudo-Lidar representation has significantly narrowed the gap between visual-based and active Lidar-based 3D object detection. However, current researches exclusively focus on pushing the accuracy improvement of Pseudo-Lidar by taking the advantage of complex and time-consuming neural networks. Seldom explore the profound characteristics of Pseudo-Lidar representation to obtain the promoting opportunities. In this paper, we dive deep into the pseudo Lidar representation and argue that the performance of 3D object detection is not fully dependent on the high precision stereo depth estimation. We demonstrate that even for the unreliable depth estimation, with proper data processing and refining, it can achieve comparable 3D object detection accuracy. With this finding, we further show the possibility that utilizing fast but inaccurate stereo matching algorithms in the Pseudo-Lidar system to achieve low latency responsiveness. In the experiments, we develop a system with a less powerful stereo matching predictor and adopt the proposed refinement schemes to improve the accuracy. The evaluation on the KITTI benchmark shows that the presented system achieves competitive accuracy to the state-of-the-art approaches with only 23 ms computing, showing it is a suitable candidate for deploying to real car-hold applications. 
\end{abstract}


\section{Introduction}
\IEEEPARstart{T}{hree} -dimensional (3D) object detection is an essential vision task in scene perception and motion prediction of autonomous driving.  Although the most powerful and reliable detection algorithms are largely developed based on Lidar scanners for its accurate scene construction \cite{zheng2021se, xu2021spg,mao2021voxel}. The over-expensive hardware cost and excessively intensive computation requirements dramatically restrict its applications in the real-world. Moreover, real-time detection is also fundamental for robotics and autonomous vehicles to avoid safety issues. However, Lidar generally perceives the surrounding environment at a rate lower than 15 HZ \cite{hesai},  
making it difficult to fulfill the demand of emergency events.

Recently, a promising alternative of Lidar with competitive cost and accuracy, called Pseudo-Lidar, was proposed in \cite{pseudolidar}. It attempts to take images as a reference to predict the 3D position of the scene, and then adopt sophisticated Lidar-based detection approaches to detect the 3D position of objects.
This research is further extended by several works in the aspects of Lidar Pseudo-Lidar fusion \cite{pl++}, coordinate transformation \cite{rethinking} and monocular application \cite{Weng_2019_ICCV_Workshops}.
However, these researches follow the route of using time-consuming Deep Neural Networks (DNNs) to estimate the depth and do not investigate the intrinsic properties of data representation in the view of Pseudo-Lidar. Although they achieve state-of-the-art 3D detection results, the serious computing latency prevents them from being deployed in practical applications. For instance, PL++:PRCNN\cite{pl++} which is the state-of-the-art Pseudo-Lidar system costing 390 ms to predict the 3D location of objects on Titan RTX GPU. It is far too slow for the timely reaction in the autonomous vehicles. 

In this paper, we explore the essential properties of Pseudo-Lidar data representation toward accurate and real-time performance. We identify the main distinguishes between Lidar data and Pseudo-Lidar data which reveal the unique profound characteristic of Pseudo-Lidar.  
We provide three observations with ablating experiments to illustrate the bottlenecks. We also introduce two point cloud data enhancement schemes, which are easy to conduct and are agnostic to different Pseudo-Lidar system.
In the first observation, we evaluate two kinds of point cloud density reduction schemes in-depth and prove their universal improvement capability to the accuracy in the detection task. Then, we introduce an easy yet effective de-smoothing strategy to augment the data representation of the point cloud generated from the stereo depth prediction, gaining the detection accuracy improvement. 
Moreover, by taking advantage of the optimization schemes mentioned above, we refine the point cloud generated from inaccurate algorithms and compare the detection results with the counterparts in which the DNNs are utilized to form the point cloud. From the ablate comparison, we found there is not necessary to deploy DNN models as depth predictors. We then provide a feasible path to construct a Pseudo-Lidar system that is practically applicable.

We evaluate our low latency Pseudo-Lidar system on both KITTI validation set and test set in comparison with several representative visual-based 3D detectors.  The proposed system achieves a 3$\times$ higher throughput than the previous fastest state-of-the-art \cite{rts3d} with an accuracy improvement up to 17\%.  When we compare the presented system with state-of-the-art detection approaches, we achieve comparable detection results with an order of magnitude faster runtime.  We also conduct ablation studies to evaluate the accuracy boost derived from different combination of data enhancement schemes proposed in this work. 

To summarize, we attribute the main contribution to three-fold:
\begin{itemize}
	\item We propose two point cloud data refinement schemes of Pseudo-Lidar to achieve the accuracy promotion for 3D detection. They are easy to adopt and agnostic to different Pseudo-Lidar detection systems.
	
	\item We demonstrate that even for low-quality depth estimation, with appropriate point cloud data enhancement, we could obtain a comparable or ever superior 3D detection result than the ones with accurate DNN-based depth estimation.
	
	\item Profit from the proposed point cloud data refinement schemes, we show the possibility of constructing the Pseuod-Lidar system with a fast but inaccurate stereo matching algorithm. We construct a lightweight 3D detection system that fully exploits prior classical algorithms to achieve competitive accuracy in comparing with the state of the arts.  More importantly, the computational latency of the proposed system is only 23 ms, showing  strong practicability.
\end{itemize}  

\begin{figure*}
	\centering
	\includegraphics[width=0.85\textwidth]{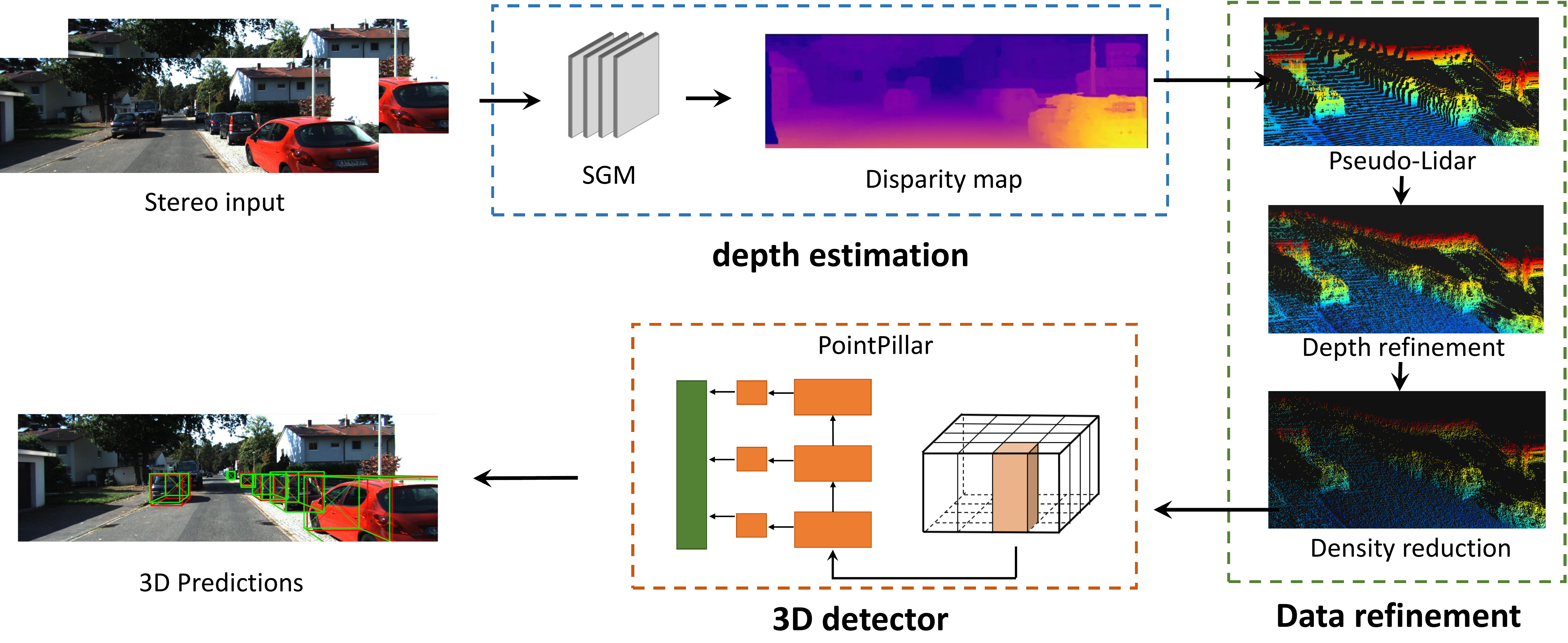}
	\caption{The overview of proposed Pseudo-Lidar 3D detection system. }  
	\label{fig:system architecture}
	\vskip - 0.3cm
\end{figure*}

\section{Related works}

The 3D object detection algorithm has wide application in many fields, such as autonomous driving and intelligent robots. It aims to determine the position, size and attitude of the object with the 3D bounding box. 
At present, there are two main methods for 3D object detection: Lidar-based detection method and stereo-based detection method.

\textbf{Lidar-based 3D object detection.} 
The 3D object detection algorithms based on Lidar mainly include three types: projection map-based, voxel-based and point-based. Typical representative  projection map-based methods MV3D\cite{chen2017multi} and AVOD\cite{ku2018joint} use the projected BEV map as the input  to conduct the 3D detection.
However, they fail to make the most of Lidar data since 
a large number of point cloud features are lost during the projection process. 
To minimize the loss of information, voxel-based approaches are first introduced in VoxelNet\cite{zhou2018voxelnet}, and further improved by SECOND\cite{yan2018second} and PointPillar\cite{pointpillar}.
The point-based  detection approaches, including 
Frustum PointNet\cite{qi2018frustum}, Point-RCNN\cite{shi2019pointrcnn}, etc.  They perform foreground and background segmentation on point cloud data and predict 3D bounding boxes through dedicated neural networks. 
Since Lidar sensors can accurately obtain the scene depth, Lidar-based methods tend to perform better than image-based methods. 
However, the expensive price of Lidar sensors has greatly incurred a hefty premium for its real-world application

\textbf{Stereo-based 3D object detection.} 
A natural solution candidate for the high cost of Lidar scanners is the highly affordable camera sensors.  
As the important progress of stereo-based 3D object detection approaches,
Pseudo-Lidar based detection systems \cite{pseudolidar,pl++,e2e} mainly utilize images to estimate the scene depth and transform it to the point cloud. Then, they predict 3D bounding boxes based on obtained point cloud data. Since the point clouds of Pseudo-Lidar share the same data representation as the Lidar, Pseudo-Lidar is compatible with existing sophisticated Lidar-based detection approaches. This allows it to directly utilize the well-studied Lidar-based detectors to gain a better detection result.
Stereo-RCNN\cite{stereorcnn} extend Faster R-CNN\cite{ren2015faster} for stereo input to detect and correlate objects in left and right images simultaneously. However, the main concern of this work is the vulnerability to the objects occlusion in the image where the dense alignment was directly operated.
Disp-RCNN\cite{disprcnn} and ZoomNet\cite{zoomnet} utilize extra instance segmentation mask to obtain the object of interest with decent improvements. RTS3D\cite{rts3d} does not require any extra information input, but it do need a extra procedure to generate latent space through a monocular-based 3D object detection algorithm. Although RTS3D\cite{rts3d} is faster than other models, the detection accuracy is not satisfactory, and it cannot be boosted with the unique data augmentation techniques of point cloud, such as sampling ground truth from the database\cite{yan2018second} and shape-aware data augmentation\cite{zheng2021se}.


\section{Approach}
	The overview of the constructed system is present in \mbox{Fig. \ref{fig:system architecture}},  We adopt SGM \cite{sgm} as the depth predictor and utilize two refinement schemes to enhance the Pseudo-Lidar data representation. Then, with the PointPillar detector \cite{pointpillar}, we could achieve solid detection in a real-time manner.
	Next, we will present in detail how we construct our system by obtaining proper Pseudo-Lidar representation to gain improvement on accuracy. 
	We first revisit the paradigm of Pseudo-Lidar based 3D detection algorithm. In particular, we investigate and analyze the distribution of the point cloud data which was largely overlooked by previous works.
	Additionally, we conduct ablation experiments to give the observations which hint us opportunities for performance promotion.
	Following these observations, we introduce a feasible path of constructing the low latency system with solid detection results.

\subsection{Rethinking the data distribution of Pseudo-Lidar}
\label{rethinking}

From the basic constitution of the Pseudo-Lidar pipeline \cite{pseudolidar,pl++},
it is clear that Pseudo-Lidar pipeline is developed by replacing the point cloud data generated from Lidar with the one from visual-based depth estimation. However, as reported in \cite{pseudolidar}, the change of input data result in a significant accuracy drop. 
To tackle this issue, a intuitive motivation is to make the point cloud data from two modalities as consistent as possible. 
Previous work \cite{pl++} introduce sparse Lidar as a reference to recorrect the point cloud data. However it is time-consuming and cost-inefficient. The work \cite{e2e} adopt an end-to-end neural network structure for the whole Pseudo-Lidar pipeline to promote the overall accuracy. However, it is less interpretability.

\begin{figure*}[htbp]
	\centering
	\subfigure[{ Direct downsampling (DD)}]{	
		\begin{minipage}[t]{0.44\linewidth}
			\centering
			\includegraphics[width=1\linewidth]{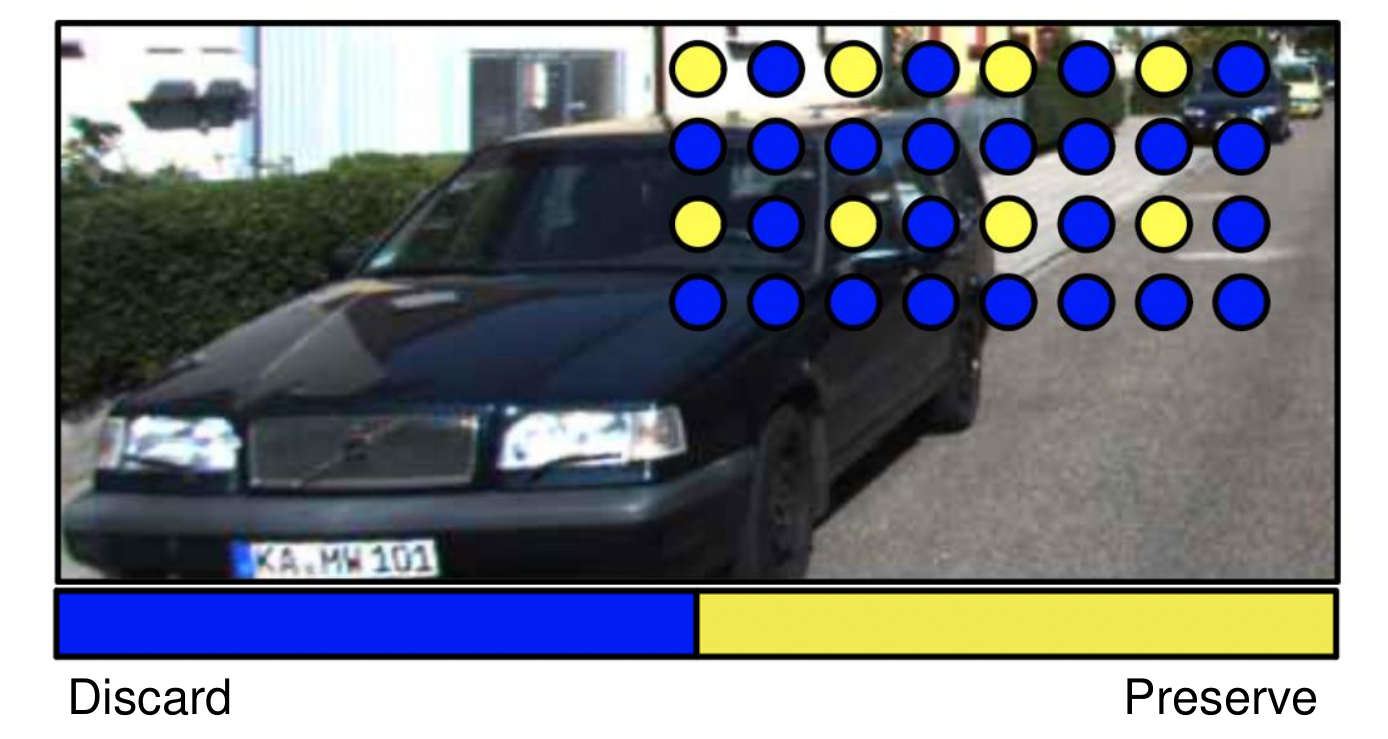}
			\label{fig:dd}
		\end{minipage}%
	}\hspace{1cm}
	\subfigure[{Adaptive downsampling (AD)}]{
		\begin{minipage}[t]{0.44\linewidth}
			\centering
			\includegraphics[width=1\linewidth]{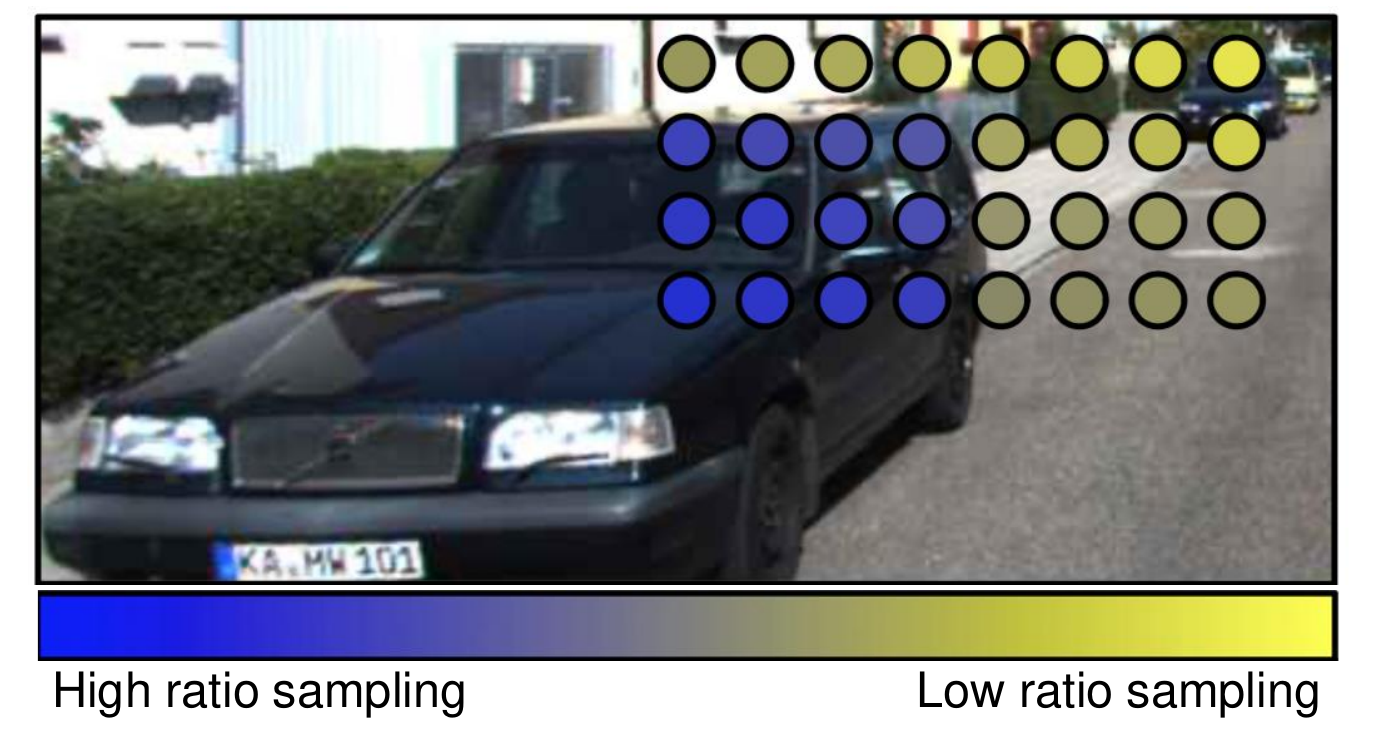}
			\label{fig:ad}
		\end{minipage}%
	}
	\centering
	\caption{ The comparison of two density reduction schemes.}
		\vskip -0.5cm
	\label{fig:sampling}
\end{figure*}

Different from these previous works, we aim to practical usage and try to resolve it in an interpretable manner.
To start with it, we first give two main distinguishes of the point cloud data from two modality sensors.
\begin{itemize}
	\item The density of data. For the nature of the high resolution of images, the point cloud in Pseudo-Lidar inherently possesses more dense data than Lidar. In fact, even for high-end Lidar with 64 channels, the total amount of valid data is only 6\% of the equivalent Pseudo-Lidar under the same scope restriction\cite{fastfusion}.
	
	\item The accuracy of the description. The depth precision of Lidar points is orders of magnitude higher than those in Pseudo-Lidar due to the principle difference. This makes the point cloud of Lidar describes the scenes more accurately. Specifically, Lidar is centimeter-level accurate for depth measurement\cite{hesai}, while the deviation of depth in Pseudo-Lidar could be as large as several meters depending on the prediction distance and the selected algorithm\cite{mccnn, stereobit}.
	
\end{itemize}

There are still some inconsistent between Lidar and Pseudo-Lidar, such as reflection data and the field of view. However, based on our experiments, they donate neglectable variables to the accuracy of the detection task. Hence, we would not go into too much discussion on these factors.

\subsection{Observations}
\label{3.2}
Based on the two distinguishes we summary in \mbox{Sec. \ref{rethinking}}, 
we investigate two point cloud data enhancement schemes accordingly. We also conduct ablate experiments and provide the \emph{observation~1} and \emph{observation~2}, respectively. Then, we follow these two observations to dive deep to investigate whether DNN-based visual depth estimation  is really necessary (\emph{observation~3}).

\textbf{Observation 1.} \emph{
	Does more dense point cloud data suggest more accurate detection performance for Pseudo-Lidar? 
}

In this observation, we shrink the density gap of the point cloud between Lidar and Pseudo-Lidar to investigate its influence on detection accuracy. 
We introduce two routines to reduce the density of the  point cloud: direct downsampling (DD) and adaptive downsampling (AD). 
To guarantee the resolution of details, we first predict the depth with original images, then downsampling the depth map to its quarter size. \mbox{Fig. \ref{fig:dd}} illustrates a simple example of direct downsampling the images.
\begin{table}
	\centering
	\begin{tabular}{cc|cc|ccc|c}
		\noalign{\global\arrayrulewidth1.2pt}
		\hline
		\noalign{\global\arrayrulewidth0.4pt}
		
		\multicolumn{1}{l}{\textbf{LEA}} & \multicolumn{1}{l|}{\textbf{PSM}} & \multicolumn{1}{l}{\textbf{PP}} & \multicolumn{1}{l|}{\textbf{PR}} & \multicolumn{1}{l}{\textbf{BL}} & \multicolumn{1}{l}{\textbf{DD}} & \multicolumn{1}{l|}{\textbf{AD}} & \multicolumn{1}{l}{\textbf{$AP_{BEV}/AP_{3D}$}}
		\rule{0pt}{13pt}\\[5pt] 
		\hline
		\checkmark                                     &                                      & \checkmark                                         &                                      & \checkmark                                     &                                  &                               & 53.55~/~43.55                              \\ 
		\checkmark                                     &                                      & \checkmark                                         &                                      &                                       & \checkmark                                &                                   & \textbf{56.96}/\textbf{46.79 }                             \\ 
		\checkmark                                     &                                      & \checkmark                                         &                                      &                                       &                                  & \checkmark                                 & 55.03~/~45.24                              \\ \hline
		\checkmark                                     &                                      &                                           & \checkmark                                    & \checkmark                                     &                                  &                                   & 47.78~/~38.97                              \\ 
		\checkmark                                     &                                      &                                           & \checkmark                                    &                                       & \checkmark                                &                                   & 49.84~/\textbf{41.48}                              \\ 
		\checkmark                                     &                                      &                                           & \checkmark                                    &                                       &                                  & \checkmark                                 & \textbf{50.30}/~41.44                              \\ \hline
		& \checkmark                                    & \checkmark                                         &                                      & \checkmark                                     &                                  &                                   & 49.46~/~39.83                              \\ 
		& \checkmark                                    & \checkmark                                         &                                      &                                       & \checkmark                                &                                   & \textbf{52.51}/\textbf{42.90}                              \\ 
		& \checkmark                                    & \checkmark                                         &                                      &                                       &                                  & \checkmark                                 & 49.07~/~39.34                              \\ \hline
		& \checkmark                                    &                                           & \checkmark                                    & \checkmark                                     &                                  &                                   & 34.34~/~31.72                              \\ 
		& \checkmark                                    &                                           & \checkmark                                    &                                       & \checkmark                                &                                   & 38.35~/~33.98                              \\ 
		& \checkmark                                    &                                           & \checkmark                                    &                                       &                                  & \checkmark                                 & \textbf{40.80}/\textbf{34.00 }                             \\ 
		\noalign{\global\arrayrulewidth1.2pt}
		\hline
		\noalign{\global\arrayrulewidth0.4pt}
		\noalign{\global\arrayrulewidth1.2pt}
		\hline
		\noalign{\global\arrayrulewidth0.4pt}
	\end{tabular}
	\vspace{5mm}
	\caption{Ablation study of the density reduction. We compare the detection accuracy of the Pseudo-Lidar pipeline that adopts three different data processing, including 1/4 direct downsampling (DD), adaptive downsampling (AD) and the default setting (BL). We combine two stereo depth estimators: LEANet (LEA) \cite{leanet} and PSMNet (PSM) \cite{psmnet}, with two 3D detectors: PointPillar (PP) \cite{pointpillar} and PRCNN (PR) \cite{shi2019pointrcnn} to cross-test and report $AP_{BEV}/AP_{3D}$ of \emph{moderate} car category at IoU=0.7.  The best result of each combination is in \textbf{bold} font.
	}
	\label{tab:sparse}
\end{table}
Concerning the adaptive downsampling, for the reason that density of point cloud decreases significantly with the increase of distance, sampling the point cloud with a fixed ratio could result in severe information loss. Therefore, we use a linear function to fit different sampling strengths according to the distances.  As displayed in \mbox{Fig. \ref{fig:ad}},  since the car and the background are located in different depths, they are adopted with different sampling ratios.
As shown in \mbox{Tab. \ref{tab:sparse}}, we compare three different settings across different depth estimators and 3D detectors combination. The results show a strong clue that all the best performances are obtained after adopting the density reduction schemes. 
It is clear that a sparser point cloud of Pseudo-Lidar could contribute to better detection accuracy. The results also show an interesting fact that PRCNN is prone to take favors from the adaptive downsampling (AD), while PointPillar achieves the best results when adopting the direct downsampling (DD) scheme.



\textbf{Observation 2.}
\emph{
	Does smoother point cloud data imply better detection performance?
}

\begin{figure*}[htbp]
	\centering
	
	\subfigure{}{
		\begin{minipage}[t]{0.475\linewidth}
			\centering
			\includegraphics[width=1\linewidth]{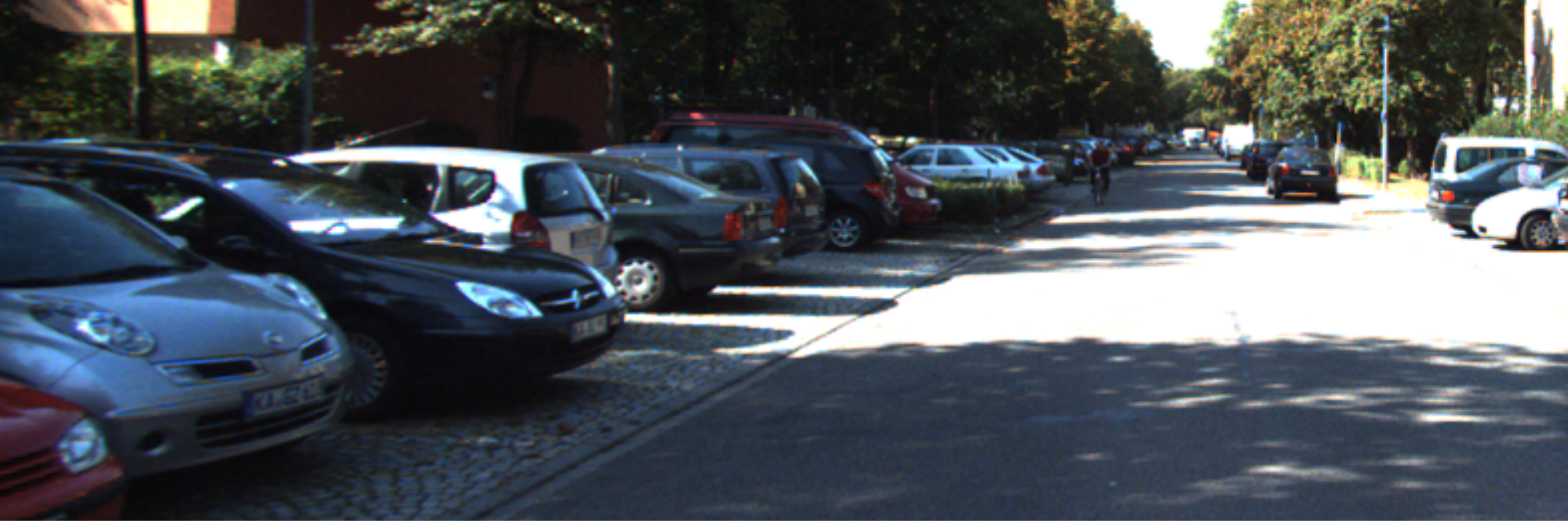}
		\end{minipage}%
	}
	\subfigure{}{
		\begin{minipage}[t]{0.475\linewidth}
			\centering
			\includegraphics[width=1\linewidth]{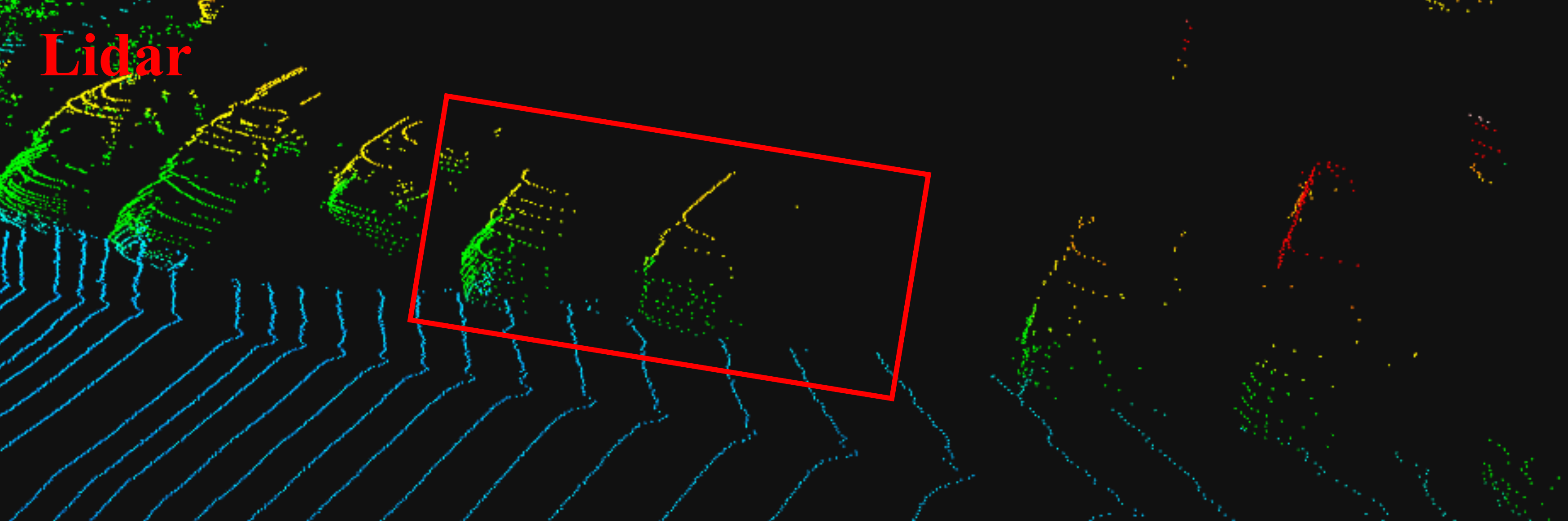}
		\end{minipage}%
	}

	\subfigure{}{
		\begin{minipage}[t]{0.23\linewidth}
			\centering
			\includegraphics[width=1\linewidth]{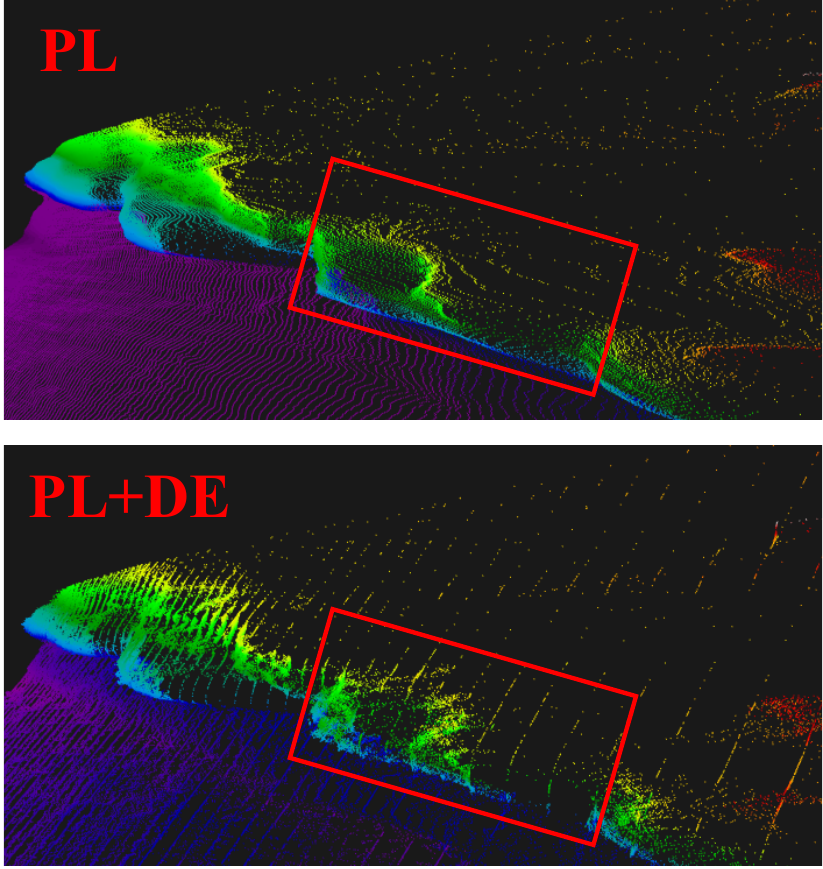}
			\caption*{\scriptsize{LEANet}}
		\end{minipage}%
	}
	\subfigure{}{
		\begin{minipage}[t]{0.23\linewidth}
			\centering
			\includegraphics[width=1\linewidth]{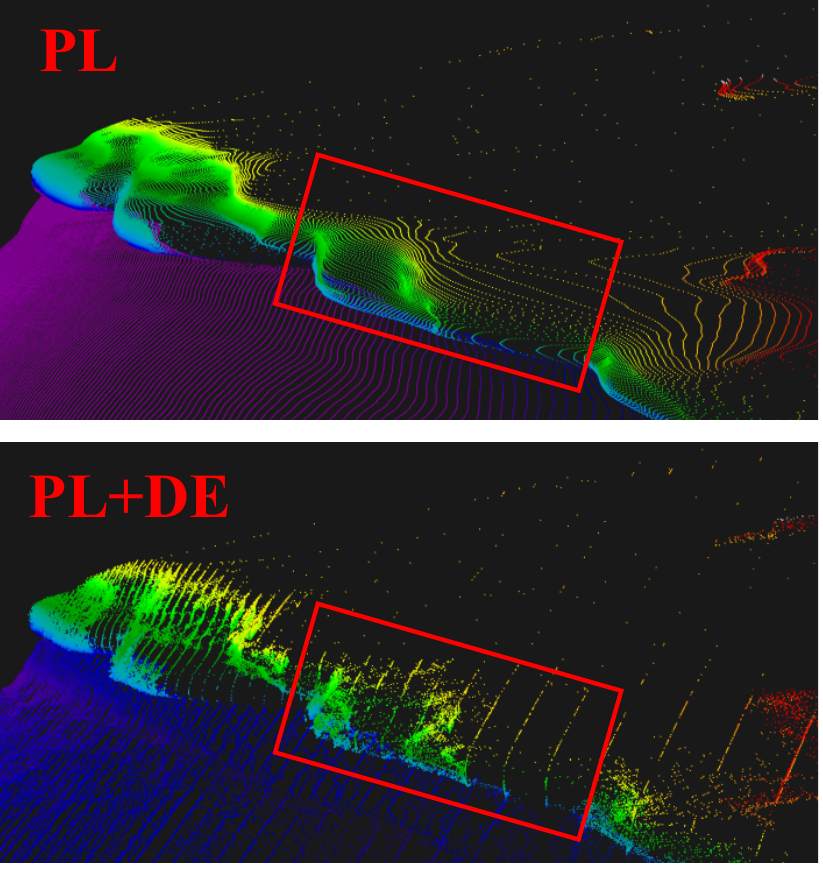}
			\caption*{\scriptsize{DECNet}}
		\end{minipage}%
	}
	\subfigure{}{
		\begin{minipage}[t]{0.229\linewidth}
			\centering
			\includegraphics[width=1\linewidth]{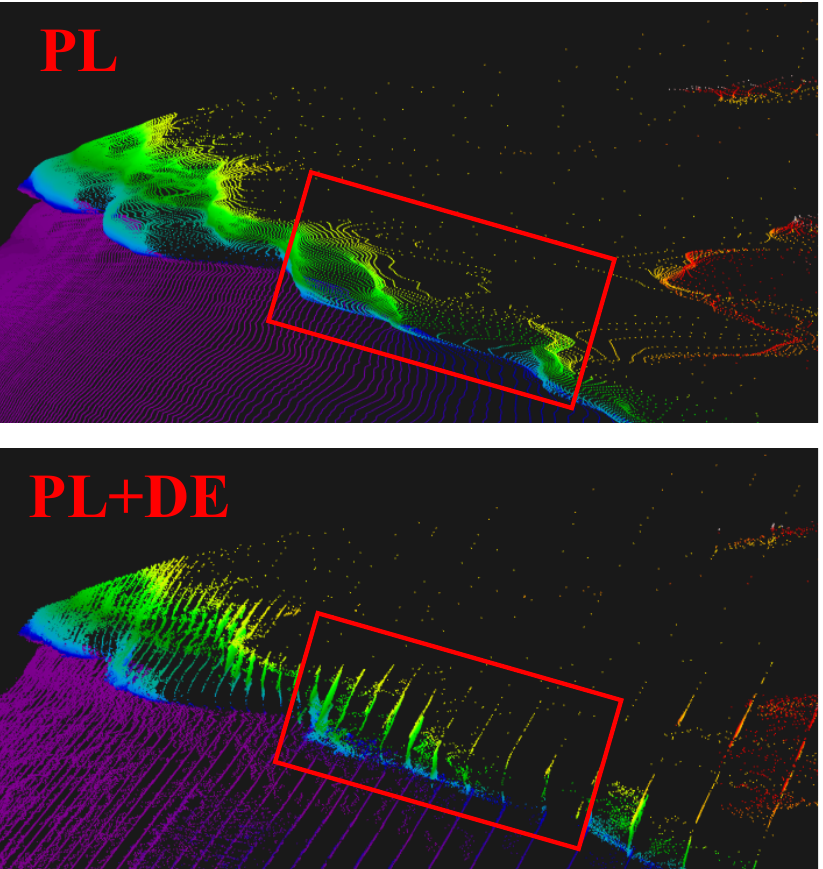}
			\caption*{\scriptsize{PSMNet}}
		\end{minipage}%
	}
	\subfigure{}{
		\begin{minipage}[t]{0.23\linewidth}
			\centering
			\includegraphics[width=1\linewidth]{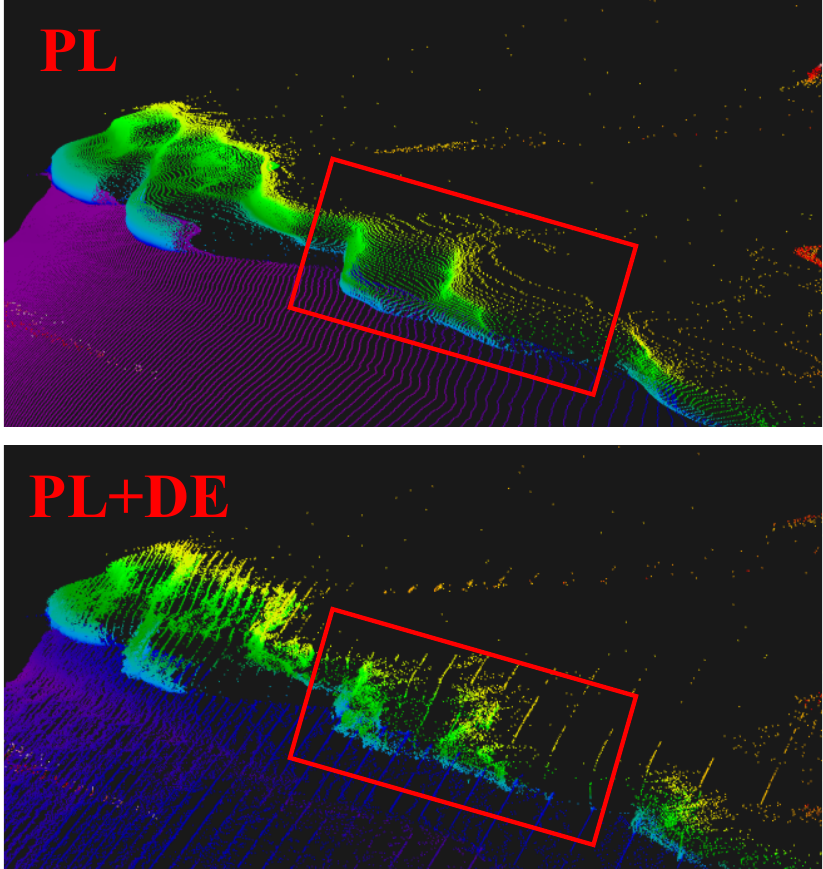}
			\caption*{\scriptsize{AANet}}
		\end{minipage}%
	}
	\centering
	\caption{ 
		Visual comparisons of the point cloud in the same perspective of a scenario.  Four representative DNN-based stereo models are illustrated. The first row is the left image and the point cloud generated from Lidar respectively. The second row is the point cloud (PL) generated from the prediction of DNN models. The third row is the refined point cloud (PL+DE).
		Strong evidence for the effectiveness of our proposed refinement scheme is how the rear of the cars becomes distinct between PL and PL+DE.}
	\label{fig:pointcloud}
\end{figure*}
In this observation, we focus on the difference in description ability of the point cloud data in Lidar and Pseudo-Lidar.
Different from the previous work \cite{pl++} that introduces extra Lidar data to correct the depth estimation in Pseudo-Lidar. 
We found the optimization opportunities from the over-smoothed depth estimation of current DNN-based models.
Since most DNN-based stereo matching algorithms for depth prediction are prone to output smooth depth maps to avoid the influence of outliers and to improve the accuracy under the visual-based depth estimation metrics. 
These smooth depth maps, however, cause shape distorting and detail loss in the 3D space. \mbox{Fig. \ref{fig:pointcloud}} illustrates the visual comparison of the point cloud generated from Lidar and DNN-based stereo depth estimator.
For this complex scene with multi cars overlapped, Lidar detailedly recovered the scenario (the right image in the first row of \mbox{Fig. \ref{fig:pointcloud}} ). As a comparison, state-of-the-art DNN-based depth predictors, such as PSMNet \cite{psmnet}, DECNet \cite{decnet}, LEANet \cite{leanet} and AANet \cite{aanet} are clearly distorting and ambiguous the  representation of the cars in the rear half of the queue (the red frame area of the second row in the \mbox{Fig. \ref{fig:pointcloud}}).

\begin{table}[]
	\centering
	\caption{\upshape Comparisons of adopting the depth refinement (DE). We report $AP_{BEV}$/$AP_{3D}$ in percentage at $IoU=0.7$. PointPillar \cite{pointpillar} was selected as the detector in this evaluation. We report Delta in \textcolor{red}{red} for the improvement with the adoption of DE and in \textcolor{blue}{blue} for the degradation.}
	\label{tab:subpixel}
	\begin{tabular}{cccc}
		\toprule
		Method   & ~~~~~~~~Easy~~~~ & ~~~~Moderate~~~~&~~~~Hard~~~~        \\ \midrule
		LEANet\cite{leanet}    & 76.65/64.38 & 53.55/43.55 & 47.79/38.43 \\
		LEANet+DE & 78.08/65.33 & 57.52/46.05 & 49.88/39.27 \\ 
		Delta &\textcolor{red}{1.43/0.95}& \textcolor{red}{3.97/2.5}& \textcolor{red}{2.09/0.84} \\\midrule
		DECNet\cite{decnet}    & 76.54/65.51 & 54.65/42.77 & 47.25/37.87 \\
		DECNet+DE & 77.52/64.26 & 57.36/45.21 & 49.68/38.95 \\ 
		Delta &\textcolor{red}{0.98}/\textcolor{blue}{-1.25}& \textcolor{red}{2.71/2.44}& \textcolor{red}{2.43/1.08} \\   \midrule
		AANet\cite{aanet}     & 76.02/65.07 & 57.25/47.19 & 51.62/41.87 \\
		AANet+DE  & 78.51/65.89 & 58.21/46.77 & 52.01/41.32 \\ 
		Delta &\textcolor{red}{2.49/0.82}& \textcolor{red}{0.96}\ \textcolor{blue}{-0.42}& \textcolor{red}{0.39/}\textcolor{blue}{-0.55} \\  \bottomrule
	\end{tabular}
\end{table}

In order to desmooth the depth map to accurately describe the scene, we utilize the photometrical information to approximately adjust the distribution of the point cloud. 
We first utilize the reference images to build up a 3D cost volume with the size $H \times W \times D$, where $H$ and $W$ are the height and the width of the image, respectively. $D$ denotes the max disparity candidate of stereo matching.  Each value of the volume $C=(x, y, d)$ represents a cost of the pixels in a window around a center pixel located in $(x,y)$ on specific disparity $d$.  With the cost volume, we fit each disparity candidate by using a quadratic curve\cite{mccnn} to increase the texture of the depth estimation. The formulation is defined as follow:
\begin{align}
	\label{eq:subpixel}
	& d_{sub} =  d_{ori} - \frac{C_	+ - C_-}{2(C_+ - 2C + C_-)}
\end{align}
where $d_{ori}$  and $d_{sub}$ are the disparities before and after the texture enhancement on the position $(x, y)$. 
$C_- = (x, y, d-1)$, $C = (x, y, d)$, and $C_+ = (x, y, d+1)$.

\mbox{Tab. \ref{tab:subpixel}} compares the performance improvement with the presence of the proposed refinement approach. 
It is remarkable that the adoption of the depth refinement scheme (DE) bring a universal performance boost for all tested stereo estimator under $AP_{BEV}$, indicating its high effectiveness. Although there are several cases of accuracy degradations under $AP_{3D}$, most stereo estimators still gain approximately 1\% performance improvement.



\begin{table*}[]
	\centering
	\caption{\upshape 
		Comparisons of different Pseudo-Lidar pipelines on 3D detection task and their corresponding stereo disparity accuracy. We report $AP_{BEV}$/$AP_{3D}$ in percentage at IoU=0.5 on KITTI validation set with PointPillar\cite{pointpillar} detector. For the accuracy metric of stereo depth estimation, We report the percentage of pixels with error bigger than 3 in the non-occluded regions (Noc) and all regions (All) in the KITTI 2015 dataset. We re-implement all compared approaches on  RTX TITAN GPU to guarantee the runtime is comparable.
	}
	\label{tab:3}
	\begin{tabular}{cccccccccc}
		\toprule
		\multirow{2}{*}{Method} & \multirow{2}{*}{Time(ms)~} & \multicolumn{2}{c}{3-pixel Err} & \multicolumn{3}{c}{IoU\textgreater{}0.5}& \multicolumn{3}{c}{IoU\textgreater{}0.7}                             \\ \cmidrule(r){3-4} \cmidrule(r){5-7} \cmidrule(r){8-10}
		&                           & ~Noc~     & ~All~~~            & ~~Easy~~        & ~~~Moderate~~~    & ~~~Hard~~~       &~~Easy~~        & ~~~Moderate~~~    & ~~~Hard~~~                                      \\ \midrule
		Lidar                   & -                        & -              & -              & 90.73/90.72 & 87.27/85.01 & 87.73/85.33    &89.16/86.61&79.61/72.72&79.07/68.90                  \\ \midrule
		PSMNet\cite{psmnet}                  & 338                          & 2.32\%        & 2.14\%         & 88.96/88.45 & 69.87/67.35 & 64.28/60.56     &74.09/61.85&49.46/39.83&42.68/35.27                \\
		AANet\cite{aanet}                   &        106                   & 2.55\%        & 2.32\%         &     90.16/89.92        &    77.85/75.87         &    72.76/69.01     &76.02/65.07&57.25/47.19&51.62/41.87                        \\
		LEANet\cite{leanet}                  &           760                & 1.65\%        & 1.51\%        & 89.64/89.33 & 74.50/71.31 & 67.66/65.48        &76.65/64.38& 53.55/43.55&47.79/38.43           \\
		DECNet\cite{decnet}                  & 43                          & 2.37\%         & 2.16\%         &     89.14/88.79        &   74.97/70.88          & 67.21/64.82   &76.54/65.51&54.65/42.77&47.25/37.87         \\ \midrule
		SGM\cite{sgm}             &1.8&8.24\%& 6.56\%&  79.94/78.27  & 64.14/59.88 & 59.34/55.15 &54.45/40.27&38.52/27.29&34.72/24.70\\
		SGM+DD+DE                    & 1.9                       &    8.31\%     & 6.62\%  & 90.02/89.83 & 78.74/77.51 & 72.58/69.52 &78.68/68.99&59.32/48.50&52.68/42.46 \\
		\bottomrule
	\end{tabular}
\end{table*}

\textbf{Observation 3.}
\emph{
	Is DNN-based stereo depth estimation really necessary for the Pseudo-Lidar pipeline?
}

3D object detection plays a central role in many applications, such as autonomous vehicles, most of which require real-time responses and high accuracy.
However, most of the current works are dedicated to improving the detection performance of Pseudo-Lidar pipelines by introducing complex architectures and algorithms, causing high computational requirements and not being capable of meeting the real-time demand.  Targeting the practical need, we investigate various Pseudo-Lidar pipelines and summarize the results as \mbox{Tab. \ref{tab:3}}. 
Comparing to the DNN-based stereo models, such as PSMNet \cite{psmnet}, AANet \cite{aanet}, LEANet \cite{leanet} and DECNet \cite{decnet}, SGM+DD+DE \cite{sgm} seems to be not comparable in accuracy of the depth estimation. However, for the task of 3D detection, 
it surpasses PSMNet \cite{psmnet}, LEANet \cite{leanet} and DECNet\cite{decnet}  by a large margin and achieves  equivalent accurate detection with AANet \cite{aanet}.
However,  it is worth noting that SGM+DD+DE only takes 1.9 ms of computing, which is 56$\times$ faster than its counterpart.
It is obviously more practical for SGM+DD+DE to be deployed in the real applications.
When we compare the SGM \cite{sgm} with its refined version: SGM+DD+DE,
we could observe a significant accuracy leap on the 3D detection task, which demonstrates that our proposed point cloud data enhancement schemes are highly effective. 
Considering the fact that SGM \cite{sgm} follows the same routines as its counterpart in the Pseudo-Lidar pipeline except for the DD and DE, which are two point cloud representation refinement schemes, we argue this can be a strong evidence to demonstrate the effectiveness of the proposed refinement schemes on improving the 3D detection accuracy.


\subsection{Accurate and Real-time Detection based on Pseudo-Lidar}

Inspired by the clues and optimization schemes introduced in \mbox{Sec. \ref{3.2}}, we construct a Pseudo-Lidar based 3D object detection system, which achieves the fastest runtime with comparable accuracy to state of the arts. As shown in \mbox{Fig. \ref{fig:system architecture}}, the system consists of three main components, depth estimation, point cloud refinement and 3D object detector. With the same paradigm of Pseudo-Lidar \cite{pseudolidar}, our system first takes the stereo pairs as input to estimate pixel-wise disparity, these disparities were then transformed to Pseudo-Lidar in the 3D world space and refined by proposed point cloud promoting schemes. In the following 3D object detection module, 3D Pseudo-Lidar were treated as real LiDAR signal to predict the 3D information of objects. 


\textbf{Depth estimation:}
Regarding the stereo depth estimation,  we take the $Observationn~3$ as a reference which indicates the impressive responsiveness and 3D detection accuracy of SGM+DD+DE. Here, the usage of depth refinement (DE) introduced in $Observation~2$ not only desmoothes the point cloud but also brings the advantage of dealing with the discontinuity of the point cloud of the SGM, making the discrete disparity values to be continuous.

\textbf{3D object Detector:} One of the biggest advantages of Pseudo-Lidar is the learnability from Lidar-based approaches when designing the detector. Therefore, instead of building the wheels from scratch, we formulate our detector on the  basis of proved approaches with the modifications. After empirical investigations and experiments, we select PointPillar \cite{pointpillar} as our 3D detector, where we could benefit from two advantages: 
(1) Unlike the classical SECOND\cite{yan2018second} encoder in which the data of point cloud are divided into small voxel containers among x, y, z dimension, pillar encoder of PointPillar \cite{pointpillar} discards the division on the z-dimension and uses several times bigger vertical column (pillar) to augment the points information. Compared to the small voxel, the bigger pillar choice increases the stability by containing more points  to make the 3D detector hard to be disturbed. Maximally eliminate the negative impact caused by noisy depth measurement.  
\textcolor{black}{We provide more experiments and analysis regarding the different pillar size choices in the \mbox{Sec. \ref{ablation}}.}
(2) The design of learning features on pillars of the point cloud enables fully 2D convolutional computing. This high-efficiency architecture is perfectly suitable for our system to meet the practical need (e.g. autonomous driving).

\begin{figure*}
	\centering
	\includegraphics[width=0.95\textwidth]{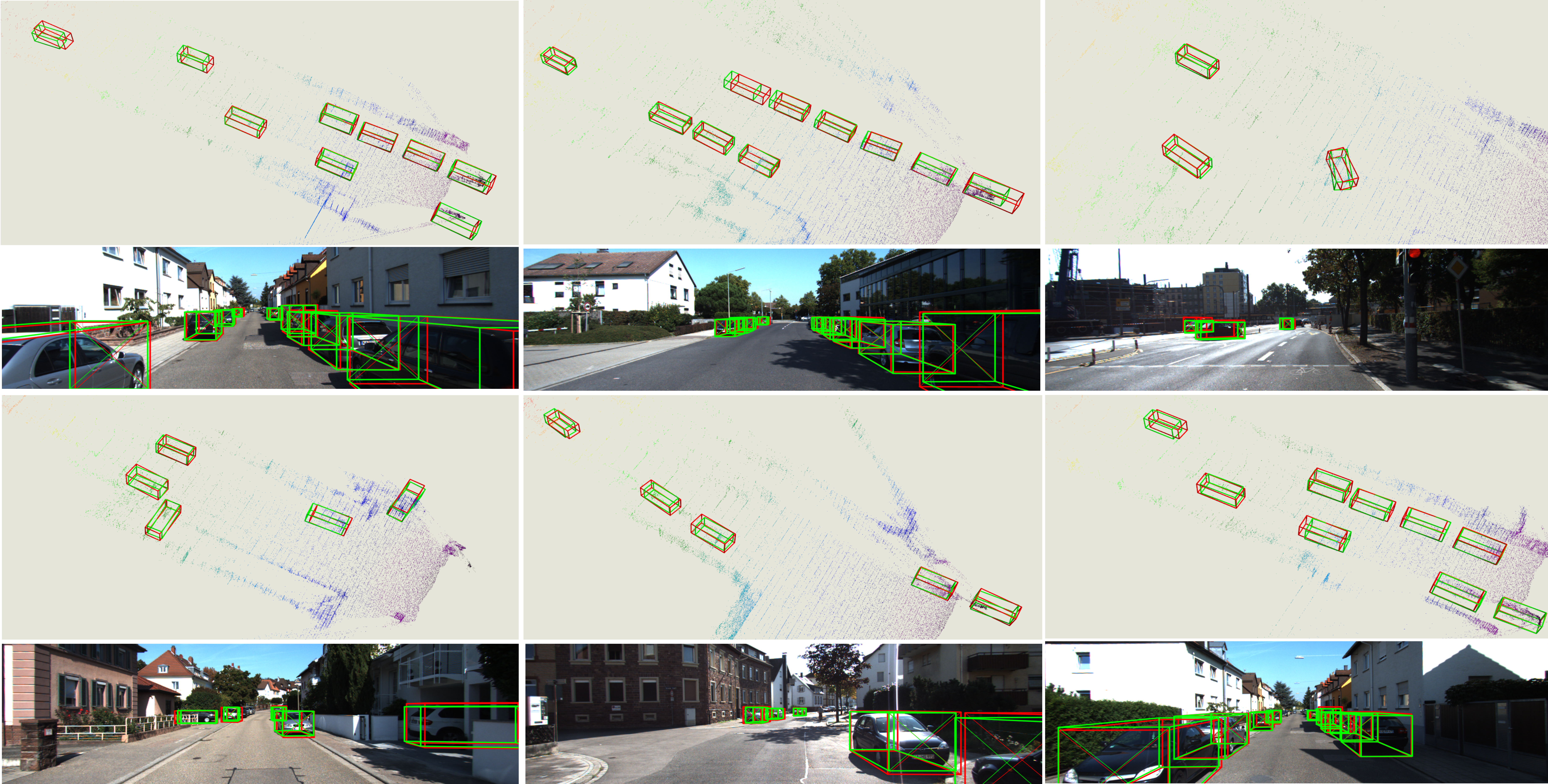}
	\caption{Qualitative comparison. We show the detection results of the proposed system on the KITTI validation set. We visualize the Ground-truth boxes in red and predicted boxes in green.  }
	\label{fig:quantitative}
\end{figure*}

\section{Experiment}
We evaluate extensive experiments of the proposed Pseudo-Lidar pipeline on both KITTI validation set and KITTI test set\cite{Menze2015CVPR} with state-of-the-art counterparts to quantitatively compare the performance. We also conduct ablation studies to investigate the optimal choice of the presence of data refinement schemes.
\subsection{Setup}
We evaluate the proposed Pseudo-Lidar pipeline on the KITTI 3D detection benchmark, which consists of 7481 images for training purposes and 7518 test images for the online benchmark. We also follow previous works \cite{chen20153d, rethinking,pseudolidar}  to divide the 7481 images into 3712 training subset and 3769 validation subset for local validation. 
For the DNN stereo models, we adopt the open-source version which is trained by authors with KITTI 2015 stereo dataset \cite{Menze2015CVPR}, Scene Flow dataset \cite{mayer2016large} and ImageNet \cite{deng2010does}. Regarding the 3D detection model, we take similar processing with \cite{pseudolidar} that train the detection model with 3712 Pseudo-Lidar data generated from stereo disparity estimation.  
In following experiments, we mainly focus on the car category and report bird's eye view ($AP_{BEV}$), 3D object detection ($AP_{3D}$) on \emph{easy}, \emph{moderate} and \emph{hard} subsets. 

\subsection{Experiment Results}

\textbf{Comparison of state of the arts on validation set.}
We compare the proposed pipeline with several state-of-the-art Pseudo-Lidar approaches. \mbox{Tab. \ref{tab:apbev}} summary the 3D object detection results under $AP_{BEV}$ and $AP_{3D}$.  
We can see the proposed pipeline is able to achieve a good balance between accuracy and runtime latency. 
It is capable of performing 3D object detection with competitive accuracy in only 23 ms computing latency. 
When we compare the proposed pipeline with RTS3D \cite{rts3d}, the fastest counterpart among state of the arts, we observe a large margin leading at $IoU>0.7$ under both $AP_{BEV}$ and $AP_{3D}$ and an impressive 4$\times$ faster runtime.  Specially, we surpass RTS3D\cite{rts3d} in the detection accuracy by at least 2.4\%, and by a maximum of 6.8\%.

\begin{table*}[]
	\centering
	\caption{\upshape Results evaluated by metric $AP_{BEV}$/ $AP_{3D}$ on validation set. Input $S$ denotes the stereo pairs, input $I$ is Instance Mask and input $C$ represents the CAD model. The best result of each column is in \textbf{bold} font. We re-implement the method with $^*$ on RTX TITAN GPU to obtain the runtime.} 
	\label{tab:apbev}
	\begin{tabular}{ccccccccc}
		\toprule
		\multirow{2}{*}{Method} & \multirow{2}{*}{Input} & \multirow{2}{*}{Runtime} & \multicolumn{3}{c}{IoU\textgreater{}0.5}                                          & \multicolumn{3}{c}{IoU\textgreater{}0.7}                                         \\ 
		\cmidrule(r){4-6}  \cmidrule(r){7-9} 
		&                              &                          & \multicolumn{1}{c}{Easy}        & \multicolumn{1}{c}{Moderate}     & Hard~        & ~Easy        & Moderate    & Hard        \\ \midrule
		
		OC-Stereo \cite{ocstereo}               & S+I          & 350 ms                    & \multicolumn{1}{c}{90.01/89.65} & \multicolumn{1}{c}{80.63/\textbf{80.03}}  & 71.06/70.34 & \multicolumn{1}{c}{77.66/64.07} & \multicolumn{1}{c}{65.95/48.34} & 51.20/40.39 \\
		ZoomNet\cite{zoomnet}                 & S+I          & -                        & \multicolumn{1}{c}{90.62/90.44} & \multicolumn{1}{c}{\textbf{88.40/79.82}}  & 71.44/70.47 & \multicolumn{1}{c}{78.68/62.96} & \multicolumn{1}{c}{\textbf{66.19/50.47}} & 57.60/43.63 \\
		Disp RCNN\cite{disprcnn}              & S+I+C      & 425 ms                    & \multicolumn{1}{c}{\textbf{90.67/90.47}} & \multicolumn{1}{c}{80.45/79.76}  & 71.03/69.71 & \multicolumn{1}{c}{77.63/64.29} & \multicolumn{1}{c}{64.38/47.73} & 50.68/40.11 \\ 
		\midrule		
		PL:AVOD$^*$\cite{pseudolidar}                 & S                   & 514 ms                   & \multicolumn{1}{c}{89.0/88.5}   & \multicolumn{1}{c}{77.5/76.4}    & 68.7/61.2   & \multicolumn{1}{c}{74.9/61.9}   & \multicolumn{1}{c}{56.8/45.3}   & 49.0/39.0    \\
		PL:PRCNN$^*$\cite{pseudolidar}               & S                   & 505 ms                          & \multicolumn{1}{c}{88.4/88.0}   & \multicolumn{1}{c}{76.6/73.7}    & 69.0/67.8  & \multicolumn{1}{c}{73.4/62.3}   & \multicolumn{1}{c}{56.0/44.9}   & 52.7/41.6  \\
		PL++:AVOD$^*$\cite{pl++}               & S                   & 399 ms                    & \multicolumn{1}{c}{89.4/89.0}   & \multicolumn{1}{c}{79.0/77.8}   & 70.1/69.1   & \multicolumn{1}{c}{77.0/63.2 }   & \multicolumn{1}{c}{63.7/46.8}   & 56.0/39.8  \\
		PL++:PRCNN$^*$\cite{pl++}             & S                   & 390 ms                    & \multicolumn{1}{c}{89.8/89.7 }   & \multicolumn{1}{c}{83.8/78.6 }    & 77.5/75.1 & 82.0/ 67.9   & \multicolumn{1}{c}{64.0/50.1}   & 57.3/45.3   \\
		E2E:PRCNN \cite{e2e} & S & 490 ms & 90.5/90.4& 84.4/79.2& \textbf{78.4/75.9}& \textbf{82.7/71.1}&65.7/51.7&\textbf{58.4/46.7}\\
		TLNet\cite{tlnet}                 & S                         & -                        & \multicolumn{1}{c}{62.46/59.5}  & \multicolumn{1}{c}{45.99/43.71} & 41.92/37.99 & \multicolumn{1}{c}{29.22/18.15} & \multicolumn{1}{c}{21.88/14.26} & 18.83/13.72 \\
		Stereo RCNN\cite{stereorcnn}             & S                         & 417 ms                    & \multicolumn{1}{c}{87.13/85.84} & \multicolumn{1}{c}{74.11/66.28}  & 58.93/ 57.24 & \multicolumn{1}{c}{68.50/54.11} & \multicolumn{1}{c}{48.30/36.69} & 41.47/31.07  \\
		RTS3D$^*$\cite{rts3d}                   & S                         & 74 ms                    & \multicolumn{1}{c}{{{90.58/90.34}}} & \multicolumn{1}{c}{80.72/79.67 }  & 71.41/70.29 & \multicolumn{1}{c}{77.50/64.76} & \multicolumn{1}{c}{58.65/46.70 } & 50.14/39.27 \\
		\midrule
		Ours                    & S                         & \textcolor{black}{\textbf{23 ms}}      & 90.36/90.15 & 79.63/78.55  & 76.47/72.97  & 80.51/69.81  & 61.11/49.36	 &56.92/46.10\\ 
		\bottomrule
	\end{tabular}
\end{table*}

E2EPL:PRCNN \cite{e2e} overall obtains the best results under both $AP_{BEV}$ and $AP_{3D}$. However, as \mbox{Tab. \ref{tab:apbev}}  illustrates, E2EPL:PRCNN \cite{e2e} achieves that at the cost of significant lower responsiveness. It takes 490 ms to propagate the end-to-end neural network and output the prediction.
In contrast, our pipeline is capable of performing decent 3D detection in 23 ms.
It is obviously more practical for the proposed pipeline to be applied in the real world. 
The qualitative examples of the proposed Pseudo-Lidar system are illustrated in \mbox{Fig. \ref{fig:quantitative}}, where the results are shown in both frontal-view images and point clouds space. Even for the cars in a distant place that are hard to identify in the image, our detection system still accurately produce the 3D bound boxes. We attribute this to the effectiveness of the proposed depth enhancement scheme (DE) which revise the depth estimation and avoid the misleading caused by over smooth.



\textbf{Comparison of state of the arts on test set.} \mbox{Tab. \ref{tab:kittitest}} summarizes the results of the car category on KITTI test set. 
Different from the results on the validation set, in the KITTI test set, the presented Pseudo-Lidar pipeline surpasses ZoomNet \cite{zoomnet}, PL++:PIXOR \cite{pl++} and Disp RCNN \cite{disprcnn} by a gap and  is only inferior to E2EPL:PRCNN \cite{e2e}. 
When we compare the results of the validation set and test set, 
we also observe a shrinking trend of accuracy gap between our Pseudo-Lidar pipeline and state of the art at $IoU>0.7$, For example, the average gap of our accuracy to the best record in each column under $AP_{3D}$ in the validation set is 1.41\%,  while in the test set, the average gap decreases to 1.18\%. 
From the perspective of practical need, our Pseudo-Lidar pipeline that achieves solid detection accuracy with only 23 ms latency is a promising solution candidate for real applications.


\begin{table}[]
	\centering
	\caption{\upshape 3D detection performance of car category on KITTI test set. We report $AP_{BEV}$/$AP_{3D}$ in percentage at $IoU>0.7$. }
	\label{tab:kittitest}
	\begin{tabular}{ccccc}
		\toprule
		Method                            & Runtime                    & Easy   & Moderate                        & Hard                             \\ \midrule
		OC-Stereo   &  350 ms & 68.8/55.1 & 51.4/37.6 & 42.9/30.2                      \\
		ZoomNet\cite{zoomnet}     &  -     & 72.9/55.9 & 54.9/38.6  & 44.1/30.9                      \\
		Disp RCNN\cite{disprcnn}  & 425 ms & 74.0/59.5 & 52.3/39.3 & 43.7/31.9                      \\ \midrule
		PL:AVOD\cite{pseudolidar}     & 514 ms & -/55.4     & -/37.1     & -/ 31.37                          \\
		PL++:AVOD\cite{pl++}   & 399 ms & 66.8/-     & 47.2/-     & 40.30/-                          \\
		PL++: PIXOR\cite{pl++} & 510 ms & 70.7/-      & 48.3/-      & 41.0/-                           \\
		E2E:PRCNN\cite{e2e} &490 ms& 79.6/64.8&58.8/43.9&52.1/38.1\\
		Stereo RCNN\cite{stereorcnn} & 417 ms & 61.6/49.2 & 43.8/34.0 & 36.4/ 28.3                      \\
		RTS3D\cite{rts3d}  & 74 ms & 72.1/58.5 & 51.7/37.3 & 43.1/ 31.1                     \\
		\midrule
		Ours        & 23 ms  & 75.0/63.3 &  54.2/42.4 & 48.1/36.4 \\
		
		%
		\bottomrule
	\end{tabular}
\end{table}

\subsection{Ablation Study}
\label{ablation}

\textbf{Impact of proposed enhancement components. }
To comprehensively investigate the contribution of the point cloud data refinement schemes in accuracy improvement, we perform an ablation experiment by enabling and disabling the point cloud reduction scheme (DD and AD) and point cloud data refinement scheme (DE).  We evaluate the effectiveness and their extra computing overhead to comprehensively evaluate the practical usability. To avoid introducing variables, we follow the default parameters and settings of PointPillar\cite{pointpillar} and retrain the detector with the point cloud data that are pre-processed with corresponding configurations.

\mbox{Tab. \ref{tab:ablation}} shows the ablation study results. Without any processing, the point cloud directly generated by the disparity map of SGM\cite{sgm} performs not satisfactory. However, once the depth enhancement (DE) is adopted, a significant accuracy boost on every metric can be obtained. Specially, the system with DE scheme in the \emph{Moderate} set  outperform 19.47 \% on $AP_{3D}$,  equivalent to a 173\% relative accuracy improvement. 
For the configuration that enables the sparing scheme (DD and AD), it is obvious that they could improve the Pseudo-Lidar system to obtain a better accuracy on every metric with approximate 1\%. 
More importantly, the extra computing overhead of proposed data refinement for point cloud are  neglectable. As shown in \mbox{Tab. \ref{tab:ablation}}, the computing latency for DE, DD and AD are 0.1 ms, 0.1 ms and 0.5 ms, respectively.  
Since these schemes are agnostic to different depth predictors and 3D detectors,  they could be embedded to different Pseudo-Lidar pipelines and result in better accuracy.

\begin{table}[]
	\centering
	\caption{\upshape Ablation study of point cloud pre-processing of our pipeline on KITTI validation set. We report $AP_{BEV}$ and $AP_{3D}$ metric at $IoU>0.7$.   }
	\label{tab:ablation}
	\begin{tabular}{cccccc}
		\toprule[0.5mm]
		\multicolumn{2}{c}{Configure}           & \multicolumn{3}{c}{$AP_{BEV}$/$AP_{3D}$}  & \multirow{2}{*}{Runtime} \\
		 \cmidrule(r){1-2}  \cmidrule(r){3-5}  
		Sparsing & DE & Easy & Moderate & Hard \\ 
		\midrule
		-        & -                         &       54.4/40.2      &        38.5/27.2         &      34.7/24.7  & 22.8 ms  \\
		-        & \checkmark &       77.9/65.5    &    57.7/46.7             &     51.6/41.1 &   22.9 ms \\
		DD       & \checkmark &     78.6/68.9        &   59.3/48.5             &   52.6/42.4    &  23.0 ms   \\
		AD      & \checkmark &    78.5/66.3        &     58.7/47.9           &  52.9/42.5   &   23.4 ms  \\
		\bottomrule[0.5mm]
	\end{tabular}
\end{table}

\begin{figure}
	\centering
	\includegraphics[trim= 100 20 60 60,clip, width=0.5\textwidth]{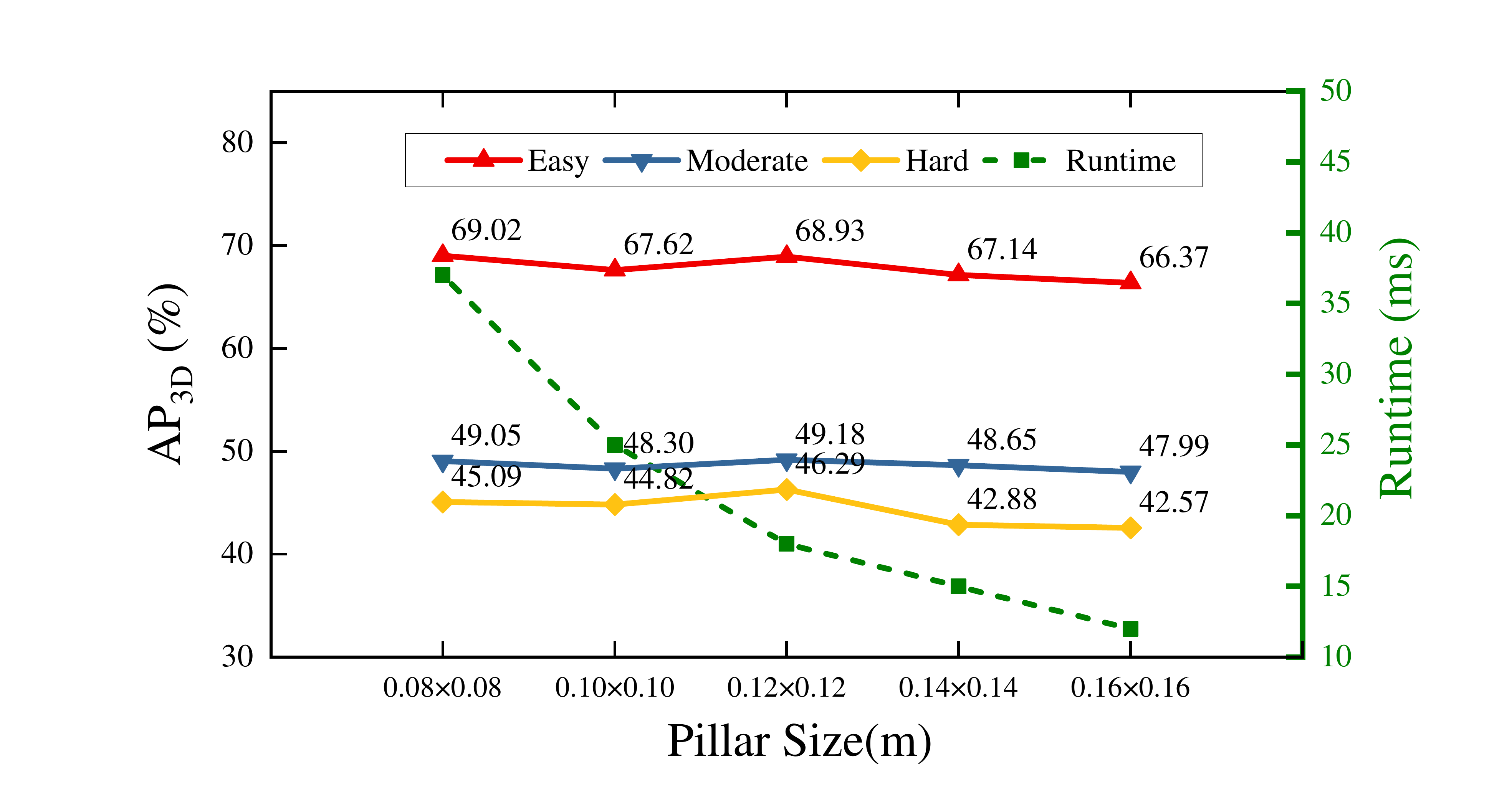}
	\caption{Ablation evaluation of pillar size. }
	\label{fig:size}
\end{figure}

\textbf{Impact of Pillar size.}
The default pillar size in PointPillar is determined according to the Lidar data. When we adopt it to the Pseudo-Lidar data, involving an adjustment could be helpful to obtain the optimal detection performance. To figure out the correspondence between the size of the pillar and the accuracy of our Pseudo-Lidar detector,  we conduct the ablation experiment for the PointPillar detector with different pillar lengths and widths (the heights of the pillar are set to $4 m$ as original PointPillar\cite{pointpillar}). Since the scope of the perspective (the size of valid point cloud data space) is fixed,  a smaller pillar size selection will lead to a bigger amount of pillars such that it could cover all data . However, more pillars mean that we need a larger tensor for computing. Therefore we also report the runtime of various pillar sizes settings.  
The comparison is illustrated in \mbox{Fig. \ref{fig:size}}, which shows a clear accuracy peak that emerges when pillar size is $0.12 m\times0.12m$.  Increasing or decreasing the pillar size will lead to an accuracy drop.  Considering the computing time is only 18 ms, we set the pillar size to $0.12 m\times0.12 m$ instead of the default setting ($0.16 m\times0.16 m$) in original PointPillar \cite{pointpillar} for a  best speed-accuracy trade-off.



\bibliographystyle{IEEEtranTIE}
\bibliography{Bibliography}



\end{document}